\documentclass[runningheads]{llncs}
\usepackage[T1]{fontenc}
\usepackage{graphicx}
\usepackage{hyperref}
\usepackage{booktabs}
\usepackage{float}
\usepackage{listings}
\usepackage{amsmath}
\usepackage{amssymb}
\usepackage{pgfplots}
\pgfplotsset{compat=1.18}
\usepackage{color}

\usepackage{todonotes}
\usepackage{zref-totpages}

\begin{document}

\title{Advancing Natural Language Formalization to First Order Logic with Fine-tuned LLMs}

\titlerunning{Natural Language to First Order Logic with LLMs}

\author{Felix Vossel\inst{1}\orcidID{0009-0005-2367-9022} \and
Till Mossakowski\inst{1}\orcidID{0000-0002-8938-5204} \and
Björn Gehrke\inst{1,2}\orcidID{0009-0007-7488-0257}}

\authorrunning{F. Vossel et al.}

\institute{Osnabrück University, Germany\\
\email{\{fvossel, till.mossakowski, bjoern.gehrke\}@uos.de}
\and University of Zurich, Switzerland}

\maketitle              
\begin{abstract}
Automating the translation of natural language to first-order logic (FOL) is crucial for knowledge representation and formal methods, yet remains challenging. We present a systematic evaluation of fine-tuned LLMs for this task, comparing architectures (encoder-decoder vs. decoder-only) and training strategies. Using the MALLS and Willow datasets, we explore techniques like vocabulary extension, predicate conditioning, and multilingual training, introducing metrics for exact match, logical equivalence, and predicate alignment. Our fine-tuned Flan-T5-XXL achieves 70\% accuracy with predicate lists, outperforming GPT-4o and even the DeepSeek-R1-0528 model with CoT reasoning ability as well as symbolic systems like ccg2lambda. Key findings show: (1) predicate availability boosts performance by 15-20\%, (2) T5 models surpass larger decoder-only LLMs, and (3) models generalize to unseen logical arguments (FOLIO dataset) without specific training. While structural logic translation proves robust, predicate extraction emerges as the main bottleneck.

\end{abstract}
\section{Introduction}

Both knowledge representation and formal specification essentially depend on formalisation of knowledge as logical theories. The formalisation process is often a bottleneck, because high-quality formalisation requires manual human effort. As a result, in many ontologies, annotations in  natural language are often much richer than the formalised logical theory. Likewise, informal requirement specifications in natural language are often much more extensive than formal specifications. This leads to a limitation of usefulness of formal methods like theorem proving, model finding and model checking, because the involved logical theory is often incomplete and only partially captures the informal natural language specification.

In order to address this problem, we aim to automate the translation from natural language to formal logic, more specifically, to first-order logic. 
While this problem has been solved to a certain extent by different methods, we aim at improving the quality of existing automated translation methods. Our approach uses large language models, which are fine-tuned and evaluated for this specific task, while being simultanesously available as general-purpose LLMs.
\section{Related Work}

\paragraph{Natural Language to First-Order Logic Translation}

Translating NL to FOL can be used to enable logical reasoning with natural language. Early approaches to this problem relied mainly on symbolic methods. They used rule-based systems with hand-crafted linguistic rules and syntactic patterns,  maping natural language structures directly to logical expressions. Pease and Murray developed an early translator from controlled English to logic for ontology-based knowledge representation \cite{pease2003english}. Bos and Markert explored logical inference techniques for recognizing textual entailment \cite{bos2005recognising}. They used model building from automated reasoning to approximate entailment relationships. Other systems often utilize frameworks like Combinatory Categorial Grammar (CCG) to build compositional semantic representations which can then be transformed to lambda expressions, FOL, or other formal languages. For example, ccg2lambda is a compositional semantics system generating logical semantic representations by combining semantic formulas bottom-up, guided by CCG parse trees \cite{martinez2016ccg2lambda}.

Barker-Plummer et al. systematically analysed the difficulties of students during a translation task. They identified, among other, the ambiguity of natural language, complex sentence structures, and uninformative automated feedback as the students key challenges in mapping natural language to formal logic \cite{barker2009dimensions}. Cai and Yates addressed the identified scalability by developing techniques for large-scale semantic parsing \cite{cai2013LargescaleSemanticParsing}. They used schema matching and lexicon extension to improve performance on database-grounded tasks. Martínez-Gómez et al. further refined compositional semantics with the ccg2lambda system \cite{martinez2016ccg2lambda}. This system produces logical semantic representations by composing formulas bottom-up from CCG parse trees.

More recently, researchers have explored methods beyond purely symbolic systems. Lu et al. demonstrated the use of Dual Reinforcement Learning (DRL) to translate NL to both Propositional and First-Order Logic \cite{lu2022ParsingNaturalLanguage}.

The emergence of large language models (LLMs) offered a new potential solution. However, LLMs often struggle with complex logical reasoning. Several recent works approach this limitations in different ways.

LogicLLaMA, a LLaMA-7B model fine-tuned specifically for NL-FOL translation, achieved performance comparable to GPT-4, but at a much lower computational cost \cite{yang2024HarnessingPowerLarge}. During their work they created the dataset MALLS, a dataset of 34K high-quality NL-FOL pairs generated using GPT-4. Other frameworks combine LLMs with symbolic reasoners. For example, LogicLM first translates NL to FOL using LLMs and then applies symbolic reasoning \cite{pan2023logiclmempoweringlargelanguage}. Experiments showed this combination performed better than using the LLM directly for reasoning. Building on top of that, Lalwani et al. presented NL2FOL \cite{lalwani2024nl2fol}, a framework to translate NL to FOL in multiple steps with few-shot prompting of an LLM. It is designed to detect logical fallacies in natural language statements. 

Thatikonda et al. focused on fine-tuning LLMs for NL to FOL translation \cite{thatikonda2024strategies}. They explored various fine-tuning strategies and data augmentation techniques. Their results show that fine-tuning significantly improves performance compared to zero-shot settings.  They also found that incrementally translating premises and conclusions of (natural language) logical arguments can improve logical inference checking.

\paragraph{Evaluation Metrics}
For the evaluation of NL to FOL translation tasks, various metrics are used in the literature. A common method is an exact, character-by-character match to evaluate whether a prediction of a model is correct \cite{levkovskyi2021GeneratingPredicateLogic,lu2022ParsingNaturalLanguage}. However, this metric is very strict as it does not account for syntactic variations in FOL expressions which do not change the underlying semantics. 

Other approaches use logical equivalence as a more robust metric. Using automated theorem provers this metric checks whether the predicted FOL expression is logically equivalent to the ground truth expression, even if they are syntactically different \cite{brunello2025EvaluatingLLMsCapabilities,han2022folio}. However, this approach either requires static signatures ensuring that the same predicates are used in both expressions or a normalization step to align predicate names.
Experiments which include NL to FOL translation only as a subtask of logical entailment use the accuracy of the entailment task as a metric \cite{thatikonda2024strategies,lalwani2024nl2fol}. 

Yang et al. defined a new metric called ``Logical Equivalence'' which is also used in other experiments~\cite{yang2024HarnessingPowerLarge,liu2025few}. In contrast to the previous methods, which rate a prediction either ``correct'' or ``incorrect'', this metric assigns a score based on the number of matching propositional valuations. However, it is questionable whether this metric is suitable for FOL, as it does not account for the complexity of quantifiers and a prediction which differs in only one valuation can still be worse than a prediction which differs in many valuations. For example, $A\wedge B\wedge C \wedge D$  is closer to $A\wedge B$ than to $\mathit{false}$, but differs from the former for $\frac{3}{16}$ of the valuations, but from latter only for $\frac{1}{16}$.

\section{Methodology}
\subsection{Model Selection}
For our formalization task, we employ both encoder-decoder and decoder-only large language models (LLMs) with different sizes to enable a systematic comparison across architectures and model sizes. Encoder-decoder models, such as T5-base~\cite{raffel2020exploring}, T5-3B~\cite{raffel2020exploring}, and Flan-T5-XXL~\cite{chung2024scaling}, are chosen based on their successful application to structured prediction tasks like text-to-text translation and semantic parsing \cite{li2023t5}, which parallel the formality and compositionality of natural language to logic translation. For the decoder-only setting, we fine-tune LLaMA3.1-8B \cite{grattafiori2024llama}, which is widely used in recent research on fine tuning LLMs for formal language and reasoning tasks \cite{yang2024HarnessingPowerLarge}, as well as Mistral-24B, recognized for its strong performance and efficiency \cite{mistral2025small3}. Additionally, we include Olmo-32B \cite{olmo20242} as a representative of modern open-source approaches in large-scale language modeling.

This selection enables an analysis of architectural influences while spanning a diverse range of parameter sizes and recent model families. Table~\ref{tab:llms} summarizes the models considered in our experiments.

\begin{table}[H]
    \centering
    \begin{tabular}{lccc}
        \toprule
        \textbf{Model} & \textbf{Architecture} & \textbf{Parameters} & \textbf{Reference} \\
        \midrule
        T5-base           & Encoder-Decoder  & 220M     & \cite{raffel2020exploring} \\
        T5-3B             & Encoder-Decoder  & 3B       & \cite{raffel2020exploring} \\
        Flan-T5-XXL       & Encoder-Decoder  & 11B      & \cite{chung2024scaling}    \\
        Olmo-32B          & Decoder-Only     & 32B      & \cite{olmo20242}           \\
        Mistral-24B       & Decoder-Only     & 24B      & \cite{mistral2025small3}   \\
        LLaMA3.1-8B       & Decoder-Only     & 8B       & \cite{grattafiori2024llama}\\
        \bottomrule
    \end{tabular}
    \caption{Large language models used for fine-tuning in our experiments.}
    \label{tab:llms}
\end{table}
\subsection{Datasets}
For our experiments, we leverage two of the largest publicly available Natural Language to First-Order Logic (NL-FOL) datasets: MALLS~\cite{yang2024HarnessingPowerLarge} and Willow~\cite{deveci2024transformer}. These datasets were selected because their considerable size makes them more suitable for training large models than, for instance, FOLIO~\cite{han2022folio}, and they contain significantly more complex natural language utterances compared to simpler resources such as Text2Log~\cite{levkovskyi2021GeneratingPredicateLogic}.

Each instance in both datasets consists of a natural language string paired with its corresponding FOL string. To ensure consistency and facilitate model training, we store the combined corpus in a unified \texttt{json} format, where each record contains both fields. For example:
\begin{center}
\begin{tabular}{ll}
\texttt{"NL":} & \texttt{"Not every student is hardworking."} \\
\texttt{"FOL":} & \texttt{"¬$\forall$x (Student(x) $\rightarrow$ Hardworking(x))"}
\end{tabular}
\end{center}

\begin{table}[ht]
    \centering
    \begin{tabular}{lrl}
        \toprule
        \textbf{Property}                & \textbf{Value} & \textbf{Description} \\
        \midrule
        Total examples                   & \textbf{49950} & NL$\rightarrow$FOL pairs in dataset \\
        Training examples                & \textbf{35964} & Size of train set \\
        Validation examples              & \textbf{3996} & Size of validation set \\
        Test examples                    & \textbf{9990} & Size of test set \\
        Unique predicates                & \textbf{62981}     & Distinct predicates occurring in FOL \\
        Unique constants                 & \textbf{2011}     & Distinct constants (entities/objects) in FOL \\
        Avg. NL length (chars)           & \textbf{86.73}     & Mean input (NL) string length [characters] \\
        Avg. FOL length (chars)          & \textbf{85.02}     & Mean FOL string length [characters] \\
        Avg. quantifiers per sample      & \textbf{1.56}     & Mean number of quantifiers per FOL entry \\
        Avg. formula depth per sample    & \textbf{6.75}     & Mean parse tree depth per FOL formula \\
        Avg. logical connectives/sample  & \textbf{3.68}     & Mean number of logical connectives ($\wedge, \vee, \neg, \rightarrow$) \\
        \bottomrule
    \end{tabular}
    \caption{Summary statistics for our unified NL-FOL dataset.}
    \label{tab:dataset_stats}
\end{table}
\subsection{Fine-tuning Procedure}
To adapt the selected pre-trained language models to the NL-to-FOL formalization task, we fine-tune each model on the combined training split described above. For the decoder-only models (LLaMA3.1-8B, Mistral-24B, and Olmo-32B), we apply Low-Rank Adaptation (LoRA)~\cite{hu2022lora} with rank $r=16$, scaling factor $\alpha=32$, a dropout rate of 0.05, and targeting all attention and feed-forward projection layers, following recommended settings for large-scale language modeling~\cite{hu2022lora}. For Flan-T5-XXL, LoRA was analogously applied to attention and feed-forward layers.  T5-base and T5-3B were trained with full fine-tuning. In all LoRa settings mixed precision training was enabled.

All models were trained with the AdamW optimizer~\cite{loshchilov2017decoupled}, a maximum of 12 epochs, early stopping (patience 4) based on validation loss, and cosine (Decoder-Only models)/linear (T5 models) learning rate scheduling with a warmup ratio of 0.05 and weight decay of 0.01. Batch sizes and learning rates are detailed in Table~\ref{tab:finetune_params}. Multi-GPU training was used for larger models as required.

All experiments were conducted on a high-performance computing cluster equipped with Intel\textsuperscript{\textregistered} Xeon\textsuperscript{\textregistered} Platinum 8480+ CPUs and either NVIDIA H100 80GB or NVIDIA A100-SXM4-80GB GPUs. The cluster and computational resources were funded by the Deutsche Forschungsgemeinschaft (DFG, German Research Foundation) – 456666331.

Except for T5-base and T5-3B (trained in \texttt{float32}), all models were fine-tuned using \texttt{bfloat16} precision. Distributed training was performed using the NCCL backend with unused parameter detection disabled. All experiments leveraged the HuggingFace \texttt{transformers}, \texttt{peft}, and \texttt{trl} libraries.

Table~\ref{tab:finetune_params} summarizes the main training hyperparameters and hardware used for each model.
\begin{table}[ht]
    \centering
    \begin{tabular}{lccccccc}
        \toprule
        \textbf{Model}    & \textbf{Arch.}  & \textbf{LoRA} & \textbf{F-FT} & \textbf{Batch Size} & \textbf{LR} & \textbf{Max Epochs} & \textbf{GPUs} \\
        \midrule
        T5-base          & Enc-Dec   & --         & \checkmark & 8  & $1 \times 10^{-3}$ & 12 & 1 \\
        T5-3B            & Enc-Dec   & --         & \checkmark & 8   & $1 \times 10^{-4}$ & 12 & 1 \\
        Flan-T5-XXL      & Enc-Dec   & \checkmark & --         & 8   & $1 \times 10^{-4}$ & 12 & 1 \\
        LLaMA3.1-8B      & Dec-Only  & \checkmark & --         & 16  & $1 \times 10^{-5}$ & 12 & 1 \\
        Mistral-24B      & Dec-Only  & \checkmark & --         & 32  & $1 \times 10^{-5}$ & 12 & 2 \\
        Olmo-32B         & Dec-Only  & \checkmark & --         & 32   & $1 \times 10^{-5}$ & 12 & 2 \\
        \bottomrule
    \end{tabular}
    \caption{Training hyperparameters and hardware for model fine-tuning. LR denotes learning rate; LoRA indicates Low-Rank Adaptation; F-FT denotes full fine-tuning.
    }
    \label{tab:finetune_params}
\end{table}
\subsection{Experiment Details}
We design a sequence of eight experiments to systematically analyze the impact of different fine-tuning strategies, data augmentations, and evaluation settings on the NL-to-FOL formalization task. Each experiment builds on the findings and methodology of the previous ones, allowing us to isolate and compare the effects of key choices in model adaptation.
\paragraph{1. Standard fine-tuning}
We fine-tune all models on pairs of natural language statements and their corresponding FOL expressions in a standard sequence-to-sequence or causal language modeling setting. For decoder-only and chat-based transformer models, we use a \texttt{ChatML}-style prompt structure: each instance is formatted into system, user, and assistant messages, with the system prompt instructing the model to output only the FOL formula (using the set $\forall, \exists, \neg, \wedge, \vee, \rightarrow, \leftrightarrow, \oplus$) and to begin with ``$\Phi=$''. For encoder–decoder (T5) models, we use a simple input–output format, prepending the source with \texttt{"translate English natural language statements into first-order logic (FOL): "} and setting the target to the corresponding FOL expression. Since T5 models rely on the SentencePiece tokenizer, they are unable to generate logical operator symbols by default. Consequently, all logical operators were substituted with their respective textual forms. The prompt templates are automatically converted to match the input requirements of each model architecture. This baseline serves as our primary reference point for all subsequent augmentations and curriculum strategies.

After fine-tuning, we save the resulting model and evaluate its performance on the test set. For decoder-only models, text generation is performed with sampling enabled and a maximum of 250 new tokens; all other generation parameters are left at their default values. For T5 models, we use beam search with five beams, a maximum output length of 256 tokens, a length penalty of 2.0, and early stopping disabled; the remaining settings follow the respective model defaults.

\paragraph{2. Token extension.}
In this experiment, we augment the model's vocabulary with dedicated tokens for all logical connectives and quantifiers used in First-Order Logic ($\forall$, $\exists$, $\neg$, $\wedge$, $\vee$, $\rightarrow$, $\leftrightarrow$, $\oplus$). Each special symbol is represented as a unique token in the tokenizer and the model's embedding matrix, ensuring they are treated as atomic units during both input encoding and output generation. This direct symbolic mapping facilitates the handling of logical structure and reduces token fragmentation of FOL expressions.

A key motivation for this experiment stems from the observation that models like T5, which rely on the SentencePiece tokenizer, cannot represent such symbols by default. In the baseline scenario, logical operators had to be replaced with textual tokens (e.g., “forall” instead of “$\forall$”), which may introduce ambiguity or inefficiency. By explicitly extending the vocabulary with the required logic symbols, we aim to assess the impact of direct symbolic encoding on formalization accuracy and model performance. Training and evaluation procedures otherwise mirror those of the baseline.
\paragraph{3. Fixed predicate list.}
In this experiment, 
for each NL–FOL pair in the dataset, the model receives an explicit, alphabetically sorted list of predicates that must appear in the FOL formalization, encouraging precise and consistent predicate usage. The list is sorted to ensure that the order of predicates does not reveal information about the logical structure of the FOL expression. This experimental setup is motivated by initial evaluation results, which indicated that one of the main challenges for large language models is correctly identifying the relevant predicates from the input. To test the hypothesis that predicate selection poses a greater difficulty than generating the correct logical structure, we provide the gold predicate list as an additional input to the model. A more detailed analysis of this issue is presented in our evaluation metrics section.
\paragraph{4. Fixed predicate list with noise.}
In this experiment, for each NL–FOL pair in the dataset, the model receives an explicit, alphabetically sorted list of predicates that must appear in the FOL formalization, but this list is augmented with additional randomly sampled predicates drawn from the predicates used in five other FOL formulas. This setup requires the model to robustly distinguish the relevant predicates among distractors when constructing the FOL expression. Such a scenario is also of practical importance, as in many downstream applications one may only be able to provide a superset of potentially relevant predicates, rather than the exact set needed for a particular formalization.
\paragraph{5. Two-step fine-tuning.}
Building on the practical consideration that gold predicate lists are rarely available in real-world scenarios, we design a two-step curriculum: In the first step, models are trained to extract the relevant predicates from the natural language input, producing a predicted predicate list. In the second step, this predicted list is provided as input to the model trained in Experiment 3 (fixed predicate list) for FOL formalization. This approach allows us to investigate whether intermediate predicate prediction can facilitate more accurate and robust logic translation when compared to direct end-to-end training.
\paragraph{6. Three-step curriculum fine-tuning.}
As an alternative to the two-step fine-tuning in Experiment 5, we address the predicate prediction challenge through a curriculum learning scheme with three stages. In the first stage, models are trained with the exact predicate list provided as input (as in Experiment 3). In the second stage, the input predicate list is augmented with noisy distractors (as in Experiment 4), increasing task difficulty. Finally, models are trained on the standard NL–FOL formalization task without any predicate hints. This progressive training approach is designed to gradually reduce the model's reliance on explicit predicate information, with the goal of improving its ability to correctly identify predicates and generate FOL expressions in a more realistic, unprompted setting.
\paragraph{7. Multilingual fine-tuning.}
In this experiment, we explore a fundamentally different approach to improving predicate understanding and generalization by leveraging multilingual training data. Specifically, we use the Mistral model and translate all natural language statements in the training set into German and French, two languages that are frequently covered in LLM pretraining alongside English. The corresponding FOL predicates, however, remain in English, serving as a consistent formal target language. Training follows the same procedure as in Experiment 1, but now includes NL–FOL pairs in three languages, while validation and test sets remain in the original English. This setting encourages the model to capture the semantic meaning of predicates independently of source language syntax, since predicates can no longer be directly mapped from the NL input. Our goal is to evaluate the cross-lingual robustness of logical formalization and to test whether multilingual exposure improves predicate abstraction and transfer.
\paragraph{8. Comparison to Other Approaches}
While previous experiments evaluated the performance of our models either internally or in comparison to other neural methods following the same approach, this section extends the analysis by comparing \textsc{LLaMA3.1-8B} to \textsc{LLaMA-2 13B}~\cite{thatikonda2024strategies} and the symbolic system \textsc{ccg2lambda}.
For the comparison with \textsc{LLaMA-2 13B}, we use the \textit{eval} subset of the FOLIO dataset, noticing that our model has not been trained on FOLIO. For the evaluation against \textsc{ccg2lambda}, we preprocessed the natural language inputs by removing all commas and appending a period to each sentence, in accordance with the input format expected by the tool.

Each experiment is designed to isolate a specific methodological or representational variable, enabling a systematic analysis of model capabilities and the underlying task. In the following, we present the detailed setup and training procedure for each experiment; results and discussion are provided in the results Section.
\subsection{Evaluation}
With all models and experimental variants in place, we systematically assess performance using three complementary metrics and robust baseline comparisons.
\paragraph{1. Exact match.}
Exact match evaluates whether the predicted and ground truth FOL expressions are identical under whitespace-insensitive string comparison, thereby abstracting away from superficial formatting differences.
\paragraph{2. Logical equivalence.}
To account for syntactically different but semantically equivalent expressions, we determine logical equivalence using an automated theorem proving (ATP) approach. Both predicted and ground truth FOL expressions are parsed using a context-free grammar that we specifically designed and implemented with the Lark parsing library~\cite{larkparser2024}. The resulting syntax trees are then transformed into Z3 format. To check for logical equivalence, we construct a Z3 solver~\cite{z3prover2024} to test the unsatisfiability of the negated bi-implication $\lnot(\varphi \leftrightarrow \psi)$, where $\varphi$ denotes the model output and $\psi$ the ground truth formula. We use the Z3 solver with a timeout of 10 seconds for each check. If the solver establishes unsatisfiability within this timeframe, the formulas are considered logically equivalent. Our approach follows the general structure proposed by~\cite{brunello2025EvaluatingLLMsCapabilities}.
\paragraph{3. Predicate matching.}
To further disentangle errors caused by incorrect predicate naming from structural or logical errors, we introduce a predicate matching step. For each predicted FOL expression, predicates are mapped to ground truth predicates using the normalized Levenshtein distance (threshold 0.6)~\cite{levenshtein1966binary}. Once predicates are aligned, both exact match and logical equivalence are recomputed on the normalized formulas. This procedure provides additional insight into whether errors are due to predicate identification or deeper logical misunderstandings.
\paragraph{Baseline comparisons.}
For Experiment~1, we first establish a baseline by evaluating each model on the dataset without any fine-tuning, in order to verify that fine-tuning indeed improves performance. We further compare the fine-tuned models to strong proprietary, open-source and reasoning models—namely, GPT-4o~\cite{openai2024gpt4o}, \textsc{DeepSeek-R1-0528}~\cite{deepseekai2025deepseekr1incentivizingreasoningcapability} with a hybrid 2500 tokens CoT setting, \textsc{DeepSeek-V3-0324}~\cite{deepseekai2024deepseekv3technicalreport} and LogicLLaMA~\cite{yang2024HarnessingPowerLarge}—for an additional performance reference. Notably, LogicLLaMA was trained on the MALLS dataset but not on Willow, providing an informative benchmark for both in-domain and out-of-domain generalization.
In addition, we evaluate the symbolic system \texttt{ccg2lambda} on the dataset to obtain a symbolic reference point for the neural baselines established in this experiment.

For Experiment~2, the results of Experiment~1 (i.e., fine-tuned model performance) are used as the baseline to assess whether the vocabulary extension yields further improvements.

For Experiments~6 and~7, Experiment~1 also serves as the baseline, allowing us to assess the relative benefits of three-step curriculum learning and multilingual training, respectively.

For Experiment~5, we use both the model’s zero-shot or pre-trained performance (without any targeted fine-tuning) and the performance from Experiment~1 as baselines, in order to capture the potential gains from intermediate predicate prediction in addition to direct fine-tuning.

For Experiments~3 and~4, we compare both to the model’s zero-shot or pre-trained performance and to the Experiment~1 baseline (i.e., standard fine-tuning). This allows us to measure not only the incremental benefit of providing explicit or noisy predicate lists, but also the overall performance gain compared to conventional fine-tuning without additional predicate information.
It is important to note that zero-shot or pre-trained performance is only evaluated for the decoder-only models, since the T5 models do not produce meaningful FOL translations without fine-tuning

In all cases, this systematic baseline setup enables robust conclusions on the effectiveness of each method and fair comparisons across experimental conditions.

\section{Results}

\subsection{Results of baselines and standard fine-tunning}

Table~\ref{tab:baseline_results} compares the accuracy of the evaluated LLMs in the baseline setup. As expected, \textsc{GPT-4o}, \textsc{DeepSeek-R1-0528}, \textsc{DeepSeek-V3-0324} and \textsc{LogicLLaMA} perform better than the other models, particularly in the few-shot setting. This can be attributed to significantly larger scale and broader pretraining corpus, as well as to the fact that \textsc{LogicLLaMA} was trained on a subset of our dataset, which may overlap with the test set. Overall, the higher scores for logical equivalence compared to exact match can likely be attributed to the model's tendency to produce non-canonical formula representations, as exemplified in Table~\ref{tab:equiv-formulas}. The results after predicate normalization are particularly insightful, as they provide initial evidence supporting our hypothesis: LLMs struggle more with correctly reproducing predicate names than with capturing the underlying logical structure.

In Table~\ref{tab:exp1_results}, we compare the accuracy of the fine-tuned LLMs in the baseline fine-tuning setup. Among the evaluated models, \textsc{Olmo-32B} performs the worst across all metrics. This may be attributed to its relatively limited pretraining, suggesting that large language models already acquire a rudimentary understanding of logical structure from pretraining on natural language alone.

If this hypothesis holds, the performance of the T5 models becomes particularly noteworthy. Despite being encoder-decoder models—unlike the decoder-only architecture of the other LLMs—they perform on par with or even outperform larger models such as \textsc{LLaMA3.1-8B} and \textsc{Mistral-24B}. This indicates that T5 models might benefit from their architecture, which explicitly encodes the input sequence before generating output. Such structural guidance during encoding could facilitate the mapping from natural language to logical forms, especially in tasks that require understanding and transformation rather than open-ended generation.
\begin{table}[ht]
\centering
\begin{tabular}{lcccc}
\toprule
\textbf{Model}        & \textbf{Exact} & \textbf{Exact (Pred.)} & \textbf{Equiv.} & \textbf{Equiv. (Pred.)} \\
\midrule
LLaMA3.1-8B*           & 2.37           & 4.10                   & 4.17            & 6.83                    \\
Mistral-24B*           & 7.24           & 10.73                   & 11.44            & 17.57                    \\
Olmo-32B*              & 1.91           & 3.44                   & 4.37            & 8.58                    \\
GPT-4o*    &  8.16           & 11.92                   & 13.02            & 20.28                    \\
LogicLLaMA*& 0.97           & 1.16                   & 3.06            & 5.27                    \\
GPT-4o**    &  10.98           & 16.60                   & 18.22            & 29.08                    \\
LogicLLaMA**& \bf 16.05           & \bf 24.79                   & \bf 18.71            &  29.45                    \\
DeepSeek-V3-0324**&  10.33           &  15.33                   &  15.53            &  23.17                    \\
DeepSeek-R1-0528****&  8.27           &  12.87                   &  18.11            & \bf 30.53                    \\
ccg2lambda***&  0.00           &  0.00                   &  0.27            &  0.43                    \\
\bottomrule
\end{tabular}
\caption{
Baseline: Model accuracy by exact and logical match, before/after predicate normalization without fine tuning. *zero-shot **few-shot *** only 8225 NL sentences could be parsed. ****CoT reasoning was used in a few-shot scenario. All scores are in \%.
}
\label{tab:baseline_results}
\end{table}
\begin{table}[H]
\centering
\begin{tabular}{lcccc}
\toprule
\textbf{Model}        & \textbf{Exact} & \textbf{Exact (Pred.)} & \textbf{Equiv.} & \textbf{Equiv. (Pred.)} \\
\midrule
T5-base               & 29.52           & 44.33                   & 32.71            & 50.17                    \\
T5-3B                 & 32.96           & 45.85                   & 36.21            & 51.54                    \\
Flan-T5-XXL           & \bf 35.79           & \bf 49.47                   & \bf 38.83            & \bf 54.79                    \\
LLaMA3.1-8B           & 28.46           & 41.87                   & 32.02            & 47.90                    \\
Mistral-24B           & 34.83           & 48.23                   & 38.18            & 53.92                    \\
Olmo-32B              & 24.29           & 36.49                   & 27.89            & 42.66                    \\
\bottomrule
\end{tabular}
\caption{
Experiment 1: Model accuracy by exact and logical match, before/after predicate normalization with standard fine tuning. All scores are in \%.
}
\label{tab:exp1_results}
\end{table}
\vspace{-1em}
\begin{table}[ht]
\centering
\label{tab:equiv-formulas}
\begin{tabular}{p{2.7cm}p{8cm}}
\toprule
\textbf{NL (Input)} & Tomatoes are red and round, while cucumbers are green and elongated. \\
\midrule
\textbf{FOL\_PRED} & $\forall x\, (\text{Tomato}(x) \rightarrow (\text{Red}(x) \land \text{Round}(x)))$\\
& $\land\ \forall y\, (\text{Cucumber}(y) \rightarrow (\text{Green}(y) \land \text{Elongated}(y)))$ \\
\midrule
\textbf{FOL\_GT} & $\forall x\, \forall y\, (\text{Tomato}(x) \rightarrow (\text{Red}(x) \land \text{Round}(x)))$\\
& $\land\ (\text{Cucumber}(y) \rightarrow (\text{Green}(y) \land \text{Elongated}(y)))$ \\
\bottomrule
\end{tabular}
\caption{Equivalent but not identical FOL translations. FOL\_PRED is the formula predicted by the \textsc{FLAN-T5-XXL} model and FOL\_GT is the ground-truth formula.}
\end{table}

In contrast, \textsc{ccg2lambda} fails to produce correct logical forms for the vast majority of inputs (\textbf{0.00\%} on not/predicate matched exact match and \textbf{0.43\% on predicate matched equivalence match}). Its performance is severely constrained by the rigidity of its manually constructed grammar, which lacks the flexibility to handle the syntactic and semantic variety present in natural language. These findings highlight the superiority of neural approaches over purely symbolic systems in the task of semantic parsing.

\subsection{Results of further experiments}
Adding explicit tokens for logical symbols yielded only marginal improvements for \textsc{Olmo-32B} and \textsc{T5-base} (less than one percentage point), while other models showed slight performance drops. This suggests that representing logical symbols using multiple subword tokens does not substantially hinder model understanding, and dedicated tokens offer no consistent performance benefit.

Models fine-tuned with ground-truth predicate names substantially outperform all other configurations. The best-performing model (\textsc{Flan-T5-XXL}) achieves \textbf{63.45\%} on the exact match and \textbf{70.27\%} on the equivalence match with only marginal improvements if predicate matching (\textbf{63.54\%} on exact match) was applied. Adding noise to predicate names leads to a slight performance drop, but the effect is marginal, confirming that the main benefit stems from access to correct predicate structure rather than stability under perturbations (Table~\ref{tab:exp4_results}).

By contrast, initial training for predicate name prediction yields considerably worse results. For example, \textsc{FLAN-T5-XXL} reaches only \textbf{25.96\%} on the exact match and \textbf{28.26\%} on the equivalence match. If predicate matching was applied \textsc{FLAN-T5-XXL} reaches \textbf{35.11\%} on the exact match and \textbf{38.95\%} on the equivalence match. This confirms our hypothesis that exact predicate alignment 
presents a hard constraint. Given the more than \textbf{60,000} unique predicates in the dataset and substantial variation in predicate usage, models struggle to generalize under these strict conditions.

The stepwise training procedure—starting with gold predicates, followed by noisy and then free-form formulas—also underperforms relative to the baseline, despite access to complete target structures during early training. \textsc{FLAN-T5-XXL} achieves just \textbf{33.83\%} on the exact match and \textbf{36.69\%} on the equivalence match, suggesting that training on ground-truth predicates might lead to overfitting on the given predicates which then reduces final performance.

Multilingual training leads to only minor degradation compared to monolingual models. \textsc{T5-3B} achieves \textbf{30.33\%} on the exact match and \textbf{32.90\%} on the equivalence match, demonstrating that multilingual inputs are compatible with consistent logical form generation and may support future cross-lingual extensions.

\begin{table}[H]
\centering
\begin{tabular}{lcccc}
\toprule
\textbf{Model}        & \textbf{Exact} & \textbf{Exact (Pred.)} & \textbf{Equiv.} & \textbf{Equiv. (Pred.)} \\
\midrule
T5-base               & 57.02           & 57.92                   & 63.22            & 64.34                    \\
T5-3B                 & 60.04           & 60.35                   & 66.28            & 66.66                    \\
Flan-T5-XXL           & \bf 62.39           & \bf 62.53                   & \bf 69.06            & \bf 69.23                    \\
LLaMA3.1-8B           & 43.76           & 43.91                   & 53.68            & 53.95                    \\
Mistral-24B           & 50.80           & 50.89                   & 59.93            & 60.07                    \\
Olmo-32B              & 36.58           & 36.90                   & 46.15            & 46.65                    \\
\bottomrule
\end{tabular}
\caption{
Experiment 4: Model accuracy by exact and logical match, before/after predicate normalization with fine-tuning on fixed predicate list with noise. All scores are in \%.
}
\label{tab:exp4_results}
\end{table}

\subsection{Results for logical arguments}

\cite{thatikonda2024strategies} argue that with traditional datasets, it is difficult to assess the quality of NL to FOL translations automatically. As an automated santiy check, they suggest to use hand-crafted sets of natural language logical arguments, whose (in)validity is known. Then, after translation to FOL, (in)validity of the argument can be checked with an automated theorem prover, and if the validity disagrees with the manually annoteted one, an error in the translation has been spotted.
FOLIO~\cite{han2022folio} is such a dataset of NL arguments.

Our model achieves competitive results (\textbf{39.70\%} valid conclusions) compared to the best version of \textsc{LLaMA-2 13B} (\textbf{37.44\%} valid conlsusions with inkremental fine-tuning and predicate verifier), even though it was not trained on the FOLIO dataset.
This suggests that the iterative fine-tuning procedure on FOLIO proposed by previous work \cite{thatikonda2024strategies} is not strictly required to attain strong performance. Additionally, our model exhibits robust behavior in multi-premise inference settings, where earlier premises are provided as contextual input—despite such scenarios not being part of the training data. This points to the model’s capacity to generalize to more complex reasoning setups without explicit supervision.

\section{Conclusion}

For the task of translating natural language into first-order logic, we have fined-tuned several large language models using the MALLS and Willow datasets, and have conducted experiments with several settings. An important aspect was the provision or learning of a list of predicate names that shall occur in the translation. While the provision of ground-truth predicate names greatly helps, learning them in a two-step process did not lead to good results.
Our fine-tuned Flan-T5-XXL model with 11 billion parameters, can outperform baselines like GPT-4o, LogicLLaMA and even DeepSeek-R1-0528, as well as a symbolic baseline. The accuracy of translations of natural language arguments from the FOLIO dataset is comparable to the baseline (note that here only (in)validity of the translated argument is considered).
The source code for our fine-tuned models is available on \href{https://github.com/fvossel/NL2FOL}{GitHub}, and the models themselves can be accessed on  \href{https://huggingface.co/collections/fvossel/nl-to-fol-685464200cad67e2cd5b0e73}{Hugging Face}.
 A common mistake in NL to FOL translation that is also present in the MALLS dataset is the confusion of hard-to-recognise Aristotelian forms: ``A pineapple is a fruit with spiky skin''
is translated to $$\exists x (Pineapple(x) \wedge  Fruit(x) \wedge  SpikySkin(x)).$$ Despite being trained with MALLS, our model correctly translates this to $$\forall x (Pineapple(x) \to (Fruit(x) \wedge HasSpikySkin(x)))),$$
which can be explained with the usage of the Willow dataset that translates such situations correctly.

We observe that reasoning strategies such as CoT can sometimes degrade performance in NL-to-FOL translation, as initially correct outputs are overwritten during the reasoning process. This aligns with the illusion of thinking phenomenon described by Apple~\cite{illusion-of-thinking}. Future work may benefit from a two-stage approach, where base model outputs are first generated and then verified or refined using a reasoning model, as suggested by Thatikonda et al~\cite{thatikonda2024strategies}.

Flan-T5-XXL achieves 55\% accuracy without a predicate list and 70\% with a noisy one, as is realistic in practice. A correlation of $-0.318$ between sentence length and exact match further suggests that shorter sentences are translated more accurately. Future work includes improving predicate extraction---e.g., splitting composite names like \texttt{ProtectFromRain}, tokenizing given predicate names, and correcting errors in the MALLS dataset to boost accuracy.

Furthermore, we plan to correct the errors in the MALLS dataset, in order to improve accuracy.

As another line of future work, we plan to cover also less  and more expressive logics used for specification and ontology development. 

\begin{credits}
\subsubsection{\ackname}
The authors want to thank Christoph Benzmüller, Dave Barker-Plummer and İbrahim Ethem Deveci for helpful discussions.

This work was supported by the Deutsche Forschungsgemeinschaft (DFG, German Research Foundation) – 456666331.

\subsubsection{\discintname}
The authors have no competing interests to declare that are relevant to the content of this article.
\end{credits}

\bibliographystyle{splncs04}
\bibliography{references}

\end{document}